\documentclass{article}

\usepackage[T1]{fontenc}
\usepackage{microtype}
\usepackage{graphicx}
\usepackage{booktabs}
\usepackage{hyperref}

\usepackage[preprint]{icml2026}

\usepackage{amsmath}
\usepackage{amssymb}
\usepackage[capitalize,noabbrev]{cleveref}
\usepackage{float}
\usepackage{caption}
\usepackage{listings}
\hypersetup{pdfsubject={Preprint}}
\graphicspath{{figures/}{../figures/}}

\icmltitlerunning{The AI Risk Observatory}

\begin{document}

\twocolumn[
  \icmltitle{The AI Risk Observatory:\texorpdfstring{\\}{ }What Can We Learn from AI Disclosures in Annual Reports\texorpdfstring{\\}{ }About Societal Resilience?}

  \icmlsetsymbol{primary}{*}
  \begin{icmlauthorlist}
    \icmlauthor{Bart Jaworski}{primary,ind}
    \icmlauthor{Andrew Sutton}{oms}
  \end{icmlauthorlist}

  \icmlaffiliation{ind}{Independent Researcher}
  \icmlaffiliation{oms}{Oxford Martin School}
  \icmlcorrespondingauthor{Bart Jaworski}{bart.jaworski.00@gmail.com}
  \icmlcorrespondingauthor{Andrew Sutton}{andrew.sutton@balliol-oxford.com}

  \icmlkeywords{AI governance, disclosure analysis, annual reports, NLP, UK corporate governance}

  \vskip 0.3in
]

\printAffiliationsAndNotice{\textsuperscript{*}Primary contributor.}

\begin{abstract}
Societal resilience research relies on access to useful and actionable data, which motivates our main research question: Can annual reports, processed at scale with LLMs, provide a useful signal about how companies disclose their response to AI? We test this by applying a reproducible two-stage classification pipeline to 9,821 annual reports from 1,362 UK listed companies (2020--2025, with partial 2026 data). We first validate the method against 474 human-annotated passages, finding high recall and moderate label-level agreement. We then report three empirical patterns: (i) between 2020 and 2025, the share of reports mentioning AI risk rose from 2.8\% to 41.2\%, while AI adoption disclosure also rose, from 13.8\% to 45.2\%, and named vendor mentions cluster around a small set of major providers led by Microsoft; (ii) disclosure varies substantially by Critical National Infrastructure sector and market segment: AIM reports disclose AI risk at far lower rates than Main Market reports, and sectors such as Energy and Data Infrastructure lag behind the rest in AI risk disclosure; and (iii) harm disclosures are near-absent (seven reports across the entire corpus). We develop a \emph{substantiveness} classification to assess the quality of the disclosure and find that most AI risk disclosure is not substantive: in 2025, 41.2\% of all reports mention AI as a risk, but only 4.3\% contain AI risk disclosure we classify as substantive.
\end{abstract}

\section{Introduction}
\label{sec:intro}

As AI systems become more prominent in corporate strategy and operations, understanding how organizations disclose AI-related risks is increasingly important. Societal resilience to advanced AI depends on knowing where AI dependencies and risks are accumulating across the economy, particularly within Critical National Infrastructure (CNI), before failures materialize. Yet measuring this at scale is difficult. Surveys are expensive and subject to response bias; incident databases capture harms after the fact; regulatory filings vary across jurisdictions and sectors. Annual reports offer an alternative: they are legally mandated, board-accountable, published alongside audited financial statements (auditors must also confirm that the strategic report is consistent with the accounts), and follow a stable year-to-year structure. Under the Companies Act 2006, the strategic report (a mandatory section of the annual report) must describe principal risks and uncertainties; the Financial Reporting Council's (non-mandatory) Guidance on the Strategic Report calls for these to be entity-specific rather than generic. This makes annual reports a potentially valuable source for tracking how listed companies understand their AI exposure.

In this paper, we explore three questions:

\begin{enumerate}
\item \textbf{Feasibility.} Can LLMs reliably extract structured AI-related signals from annual reports at scale?
\item \textbf{Findings.} What patterns emerge when we apply this method to the near-complete population of UK-incorporated listed companies from 2020 to 2025 (with partial 2026 data)?
\item \textbf{Signal quality.} How useful and informative are these disclosures: are they substantive and specific, or largely boilerplate?
\end{enumerate}

In terms of feasibility, we validated a two-stage LLM classification pipeline against 474 human-annotated passages drawn from 30 reports across 10 of the 12 CNI sectors in our taxonomy, finding high recall and moderate label-level agreement. In terms of findings, we document a rapid rise in AI risk, adoption, and vendor disclosure between 2020 and 2025, substantial variation between CNI sectors and market segments (i.e., Main Market vs.\ AIM), and a near-total absence of harm disclosures. In terms of signal quality, we find that most disclosures are not substantive, which limits the conclusions that can be drawn from this data source alone.

The primary contribution is methodological: we demonstrate that LLM-based extraction from annual reports is feasible and reproducible, and that it produces trend signals consistent with widely reported developments in AI adoption. A secondary contribution is empirical: we provide the first large-scale, CNI-decomposed view of AI disclosure across UK listed companies, covering Main Market, AIM, and Aquis issuers. Additionally, we create a mapping of companies in our dataset onto the CNI sectors using a set of LLM autograders.\footnote{Since not all companies operate in or contribute to the CNI sectors, 970 of the 1,362 companies with a processed report are mapped to a primary CNI sector, while the remaining 392 are classified under the ``other'' category.}

\section{Related Work}
\label{sec:related}

\textbf{Financial disclosure NLP.}
\citet{loughran2011liability} established annual reports as a productive natural language processing (NLP) corpus and documented that general-purpose dictionaries misclassify financial language; their 2016 survey \citep{loughran2016textual} reviews the NLP methods most commonly applied to financial text, including dictionary-based classification, readability metrics, and document similarity, establishing best practices that inform our pipeline design. \citet{lang2015textual} demonstrate that large-sample textual analysis of non-US annual reports, across 42 countries including the UK, is feasible and informative. More recent work applies LLMs to disclosure text: \citet{park2025llm} uses LLMs and LLM agents to quantify material physical risks from textual risk disclosures in financial statements.

\textbf{Substantiveness and boilerplate.}
\citet{dyer2017evolution} show systematic inflation in US 10-K reports, which grow longer while becoming more redundant, motivating the question of whether AI disclosure follows the same pattern. \citet{brown2011large} show that year-to-year changes in \emph{Management Discussion and Analysis (MD\&A)} text can be used as a proxy for how much new information companies disclose.

\textbf{Disclosure and governance of AI.}
The most comparable study, \citet{ubertibona2026ai}, analyzes over 30,000 US filings and documents steep growth in AI risk mentions alongside persistently generic mitigation detail, consistent in direction with our UK-focused findings. \citet{bonson2023algorithmic} find algorithmic decision-making disclosure in Western Europe largely voluntary and uneven. \citet{chiu2025generative} raises the concern that generative AI can itself be used in the writing of narrative disclosures, altering the authorship and reliability of the text being analyzed.

\textbf{Observatory-style monitoring.}
Domain-adapted language models such as ClimateBERT \citep{webersinke2022climatebert} make thematic classification of corporate disclosure text tractable, and \citet{ferjancic2024esg} apply textual analysis to corporate sustainability reporting to relate disclosed themes to firm-level ESG outcomes. We apply the same logic longitudinally to AI-related disclosure.

\section{Data and Methodology}
\label{sec:method}

\subsection{Universe and Corpus}

The target universe is UK-incorporated entities listed on the London Stock Exchange (Main Market and AIM) and the Aquis Stock Exchange (AQSE). After removing 191 companies incorporated outside the UK, the universe contains \textbf{1,469 companies}. The analysis window covers \textbf{publication years 2020--2025}, with partial 2026 data (339 reports) included in trend visualizations and reported separately, but not used for year-on-year comparisons. Of the 1,469 companies in the universe, 107 had no machine-readable annual report available across any analysis year (typically smaller AIM or Aquis issuers for which neither the data provider nor the Companies House route described below yielded a usable document), leaving \textbf{1,362 companies} with at least one processed report. The full corpus contains \textbf{9,821 processed reports} (\cref{tab:corpus}): 9,482 for 2020--2025 plus the 339 partial-2026 reports. The unit of analysis is a filed annual-report document rather than a company-year: a company can contribute more than one document in a publication year (for example, an ESEF package and a PDF version of the same report, multi-part uploads, or copies filed on a second exchange), so the 2025 corpus comprises 1,561 documents from 884 companies (739, 896, 1,020, 1,002, 1,032, 884, and 207 companies for 2020--2026 respectively), and all ``\% of reports'' rates below are document-weighted. Company-level rates are close to document-level rates for the headline series (in 2025: any AI mention 65.6\% vs.\ 65.5\%, AI risk 41.6\% vs.\ 41.2\%, adoption 46.7\% vs.\ 45.2\%), but differ more for the partial 2026 data (AI risk 72.0\% vs.\ 66.4\%).

\begin{table}[t]
\caption{Corpus and validation summary.}
\label{tab:corpus}
\vskip 0.1in
\centering
\small
\begin{tabular}{@{}lr@{}}
\toprule
Item & Value \\
\midrule
Companies in universe & 1,469 \\
Companies with $\geq$1 report & 1,362 \\
Processed reports & 9,821 \\
Analysis years & 2020--2025 (+2026 partial) \\
Validation passages & 474 \\
Validation companies & 15 \\
CNI sectors in validation & 10/12 \\
Classification model & Gemini 3 Flash \\
\bottomrule
\end{tabular}
\end{table}

Annual reports were sourced from a third-party data provider\footnote{We used data from FinancialReports.eu, a commercial provider that supplies annual reports and other regulatory filings in markdown format, which is considerably easier to process than raw PDFs.} that delivers annual reports in markdown format with structured metadata. This approach was adopted after initial experiments with direct PDF retrieval from the Companies House API followed by OCR conversion, which proved noisy, expensive, and slow at full-corpus scale. The third-party data source provides approximately 95\% coverage for FTSE\,350 companies but only around 28\% for AIM, reflecting lighter electronic filing requirements among smaller issuers. For company-years missing from the provider, we fell back on the Companies House route at much smaller scale: raw PDFs were retrieved from Companies House and converted into markdown using Mistral's OCR batch API.

\subsection{Preprocessing}
\label{sec:preprocessing}

First, the markdown reports are normalized, and structural metadata (headings, section boundaries) are preserved. Then, an \textbf{AI keyword gate} is applied before any LLM classification: a passage must contain at least one term from a fixed list of core AI terms (\emph{artificial intelligence}, \emph{machine learning}, \emph{large language models}, \emph{generative AI}, and clearly AI-specific techniques such as neural networks, computer vision, or natural language processing), named AI products and vendors, or AI application terms (e.g., robotic process automation, predictive analytics, chatbot).\footnote{The complete keyword list is provided in \cref{app:keywords}.} Generic terms such as ``data analytics,'' ``digital transformation,'' and ``automation'' are not triggers on their own. Because some application terms are broader than AI, passages that do not explicitly concern AI are removed by the Stage~1 explicitness rule described below; the gate itself is conservative in that it never admits a passage without one of the listed terms.

Passages that pass through the gate are extracted with a context window that spans two paragraphs before and after the trigger text. This yields 24,189 passages (hereafter \emph{AI mentions}), each preserving the surrounding context and linked structured metadata, including company identity, publication year, market segment, CNI sector, and report section.

\subsection{Two-Stage Classification Pipeline}
\label{sec:pipeline}

\textbf{Stage~1} assigns a high-level mention type. Each AI mention receives one or more labels: \textsc{adoption} (current, company-specific AI deployment), \textsc{risk} (a downside or exposure attributed to AI), \textsc{vendor} (explicit third-party AI provider), \textsc{harm} (past, specific AI-caused incident), \textsc{general/ambiguous} (AI mentioned but not meeting the criteria for adoption, risk, vendor, or harm), or \textsc{none} (keyword-gate false positive: the trigger term does not actually refer to AI). At the passage level, labels are non-mutually exclusive with two exceptions: \textsc{none} excludes all other labels, and \textsc{general/ambiguous} is assigned only when no other label applies, so it cannot co-occur with \textsc{adoption}, \textsc{risk}, \textsc{vendor}, or \textsc{harm}. At the report level, a report may contain both kinds of passages. Stage~1 uses conservative labeling criteria to limit false positives, given the high volume of generic statements about AI in annual reports: adoption requires first-person company language indicating actual deployment, and risk requires that a downside or exposure be attributed to AI rather than merely mentioned near it; Stage~2 then grades how explicit that attribution is. The classifier also reports a confidence score (0--1) for each label, and we retain labels with confidence $\geq 0.2$, so conservatism comes from the labeling rubric and the keyword gate rather than from a strict confidence cut-off.\footnote{The design of the classification prompts is summarized in \cref{app:prompts}; the full prompt text is reproduced in \cref{app:prompt_text}.}

\textbf{Stage~2} applies three further classifiers to Stage~1 AI adoption, risk, and vendor mentions:

\emph{Adoption classifier.} Scores each mention across three non-mutually exclusive types: \textsc{non-llm} (traditional ML, computer vision, predictive analytics), \textsc{llm} (large language models and generative AI), and \textsc{agentic} (autonomous AI systems executing tasks without continuous human oversight). Adoption-type scores are computed by the pipeline but are not analyzed in this paper; their precision for rare types such as agentic adoption has not been validated (see \cref{sec:limitations}).

\emph{Risk classifier.} Maps AI risk to one or more of ten categories: Strategic/Competitive, Cybersecurity, Operational/Technical, Regulatory/Compliance, Reputational/Ethical, Third-Party/Supply Chain, Information Integrity, Workforce Impacts, Environmental Impact, and National Security (full category definitions are provided in \cref{app:risk_taxonomy}). Each risk mention also receives a 1--3 attribution signal: 3 means explicit causal language linking the risk to AI in the same statement, 2 means clear but inferential attribution, and 1 means indirect or weak attribution (\cref{app:prompts}); we call a risk mention \emph{explicit} when its attribution signal is 3. We developed this taxonomy from first principles and stress-tested it via an open-taxonomy classification test run on 100 AI risk mentions (96 classified successfully): the model produced 107 distinct labels that collapsed, after manual consolidation, into broad themes, none of which fell outside the fixed taxonomy.\footnote{Details of the open-taxonomy validation are provided in \cref{app:taxonomy_validation}.}

\emph{Vendor classifier.} Identifies the AI provider referenced (Microsoft, Google, OpenAI, Amazon, Nvidia, Meta, and others), plus four special categories: \textsc{internal} (in-house systems), \textsc{other} (a named provider outside the tag set), \textsc{undisclosed} (the passage indicates reliance on a third-party AI provider without naming it), and \textsc{open-source model} (an open-weight model used without a named provider).

All three classifiers assign a three-level \textbf{substantiveness rating}: \emph{1: boilerplate} (generic; could appear unchanged in any report), \emph{2: moderate} (identifies a specific area but lacks mechanism or mitigation detail), or \emph{3: substantive} (describes a specific mechanism and provides concrete controls, named systems, or measurable targets). For risk, the distinction is concrete: a \emph{boilerplate} passage such as ``AI is a rapidly evolving technology that may impact our business'' could appear unchanged in almost any report; a \emph{moderate} passage such as ``AI regulation may affect our compliance obligations'' identifies a risk area but gives the reader no mechanism or mitigation; a \emph{substantive} passage such as ``We allocated \pounds5M to bring three high-risk AI systems into conformity with the EU AI Act by Q3~2025'' provides both. For report-level statistics, a report is assigned, for each classifier, the substantiveness band held by the largest number of its passages (ties resolved to the higher band); every report-level substantiveness figure in this paper, including the per-category and per-vendor splits in \cref{fig:vendor_landscape} and \cref{fig:risk_subst}, uses this rule.

\subsection{Validation}
\label{sec:validation}

The pipeline was validated against 474 human-annotated AI mentions drawn from 30 reports across 15 companies, covering 10 of the 12 CNI sectors in our taxonomy (no Data Infrastructure or Civil Nuclear company is included). Agreement between LLM and reference labels, measured by mean Jaccard similarity over label sets, is 0.73 for Stage-1 mention types, 0.41 for Stage-2 adoption types, and 0.33 for the ten-category risk taxonomy (mean 0.49, median 0.41 across the three multi-label categories). Per-label precision and recall for the Stage-1 labels are reported in \cref{tab:validation}: recall is consistently high (0.91--0.97) while precision is lower and label-dependent, and manual inspection of low-agreement multi-label categories (especially risk) suggests that much of the gap reflects higher LLM label completeness: the classifier tends to assign more, and frequently valid, labels per passage than the annotator initially did. Stage-2 vendor tags could not be scored at label granularity because the archived validation output records only a binary vendor flag; vendor presence is instead validated as a Stage-1 label (precision 0.67, recall 0.91). The harm label is so rare that the validation set contains no true instance, and the classifier's 40 false positives on this label motivated the corpus-wide manual review of all harm-tagged passages (\cref{sec:harms}). The reference labels were produced by a single annotator and subsequently adjudicated: every disagreement between the initial human labels and the classifier output was manually re-reviewed and resolved to the label set judged correct. Because no second independent annotator exists, no inter-annotator agreement baseline is available. Coverage limitations of the validation set are discussed in \cref{sec:limitations}.

\begin{table}[t]
\caption{Per-label precision and recall of the Stage-1 classifier against the adjudicated validation set (474 passages; support is the number of reference-positive passages per label; labels are non-mutually exclusive). The validation set contains no true harm instance: all 40 harm-tagged passages were false positives.}
\label{tab:validation}
\vskip 0.1in
\centering
\small
\begin{tabular}{@{}lrrr@{}}
\toprule
Stage-1 label & Support & Precision & Recall \\
\midrule
adoption & 147 & 0.57 & 0.97 \\
risk & 101 & 0.82 & 0.97 \\
vendor & 33 & 0.67 & 0.91 \\
general/ambiguous & 159 & 0.58 & 0.96 \\
harm & 0 & 0.00 & --- \\
none & 165 & 0.93 & 0.97 \\
\bottomrule
\end{tabular}
\end{table}

This validation provides evidence on the feasibility question: LLM-based extraction of structured AI disclosure signals from annual reports is feasible, with high recall and moderate precision at the mention-type level where the headline trends are computed; finer-grained labels (the risk taxonomy and Stage-2 categories) carry more noise, and we interpret them accordingly. The method has known limitations (discussed in \cref{sec:limitations}) but is sufficient to support the trend-level analysis that follows.

\section{Results}
\label{sec:results}

The feasibility question was addressed by the validation in \cref{sec:validation}. Sections~\ref{sec:surge}--\ref{sec:uneven} address the findings question, and \cref{sec:quality} addresses the signal-quality question.

\subsection{AI Mentions Have Risen Rapidly}
\label{sec:surge}

The most prominent pattern we find is the scale and speed of AI mention growth. In 2020, 199 of 1,007 processed reports (19.8\%) contained at least one AI mention. By 2025, that figure had risen to 1,023 out of 1,561 reports (65.5\%). In partial data for 2026, the rate reaches 77.3\%. AI risk mentions grew even faster from a smaller base: the share of reports mentioning AI risk rose from 2.8\% (28 reports) in 2020 to 41.2\% (643 reports) in 2025, a roughly fifteen-fold increase in share (the number of such reports rose about 23-fold as the corpus grew), and partial 2026 data (339 reports) suggest the rise is continuing, though these figures are not final. The complete year-by-year breakdown is in \cref{tab:prevalence} (\cref{app:prevalence}).

\begin{figure}[tbp]
\centering
\includegraphics[width=\columnwidth]{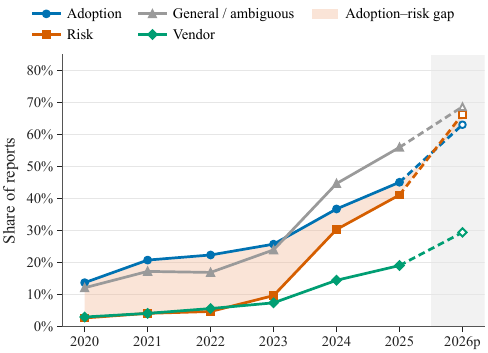}
\caption{Annual disclosure rates for AI adoption mentions, AI risk mentions, general/ambiguous AI mentions, and AI vendor mentions, 2020--2026 (2026 is partial). The shaded region shows the difference between the AI adoption-mention share and the AI risk-mention share.}
\label{fig:trends}
\end{figure}

Three phases are visible (\cref{fig:trends}). \textbf{2020--2022}: adoption mentions grew steadily (13.8\%\,$\to$\,22.5\%) while risk mentions lagged (2.8\%\,$\to$\,4.8\%), and the difference between the adoption-mention rate and the risk-mention rate widened to 17.7 percentage points (pp). \textbf{2023}: risk mentions first rose sharply, more than doubling (4.8\%\,$\to$\,9.8\%) in reports drafted in the months after ChatGPT's release, although most of these reports cover fiscal years ending in 2022. \textbf{2024--2025}: the step change came in 2024, the first full reporting cycle after ChatGPT, with the risk-mention rate tripling in a single year (9.8\%\,$\to$\,30.4\%) and reaching 41.2\% by 2025, nearly at parity with adoption (45.2\%).

The general/ambiguous category (AI explicitly mentioned but without meeting the threshold for adoption, risk, vendor, or harm) increased from 12.2\% of reports in 2020 to 56.1\% in 2025, overtaking adoption as the largest type of mention. Restricting to reports whose \emph{only} AI mentions are general/ambiguous, the 2025 rate is 6.9\% (108 of 1,561 reports; \cref{fig:composition} in the appendix shows the mutually exclusive composition over time). This suggests that mentioning AI has become common in annual reports, but that generic AI language is mostly layered on top of, rather than substituting for, more specific disclosure.

Despite these trends, 918 of 1,561 annual reports (58.8\%) in 2025 still did not mention the risk of AI at all.

\subsection{Disclosed Risk Categories Are Diversifying}
\label{sec:diversification}

In 2025, companies mentioned a broader range of AI risk categories than competitive concerns alone. Strategic/Competitive remains the leading category (27.4\% of all 2025 reports), but Cybersecurity has risen to a close second (25.7\%), narrowly ahead of Operational/Technical (24.2\%); cybersecurity risk sections commonly cite AI-enabled phishing, adversarial attacks, and AI-accelerated fraud.

Two categories show exceptional growth from near-zero bases. \textbf{Information Integrity} (AI-generated misinformation, deepfakes) rose from 1~report (0.1\%) in 2020 to 162 (10.4\%) in 2025, and \textbf{Workforce Impacts} from 3 (0.3\%) to 132 (8.5\%), a roughly 28-fold rise in share. Their emergence in corporate reporting suggests a broadening of the disclosed risk landscape from competitive and technical concerns to societal and human-capital themes. \Cref{fig:risk_heatmap} plots the complete time series for all ten risk categories; \cref{fig:risk_subst} in the appendix compares the prevalence of categories under two definitions: all mentions and only substantive mentions.

\begin{figure}[tbp]
\centering
\includegraphics[width=\columnwidth]{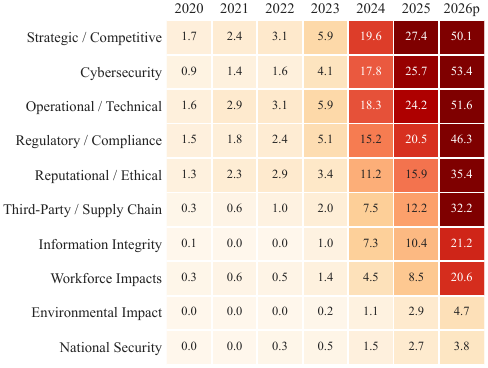}
\caption{AI risk category prevalence by publication year (\% of reports), 2020--2026 (2026 partial). Rows are ordered by 2025 prevalence. The color scale is set on the complete years 2020--2025; 2026p values exceed the scale and are shown for reference. Categories such as Information Integrity and Workforce Impacts emerge as frequently disclosed only from 2024.}
\label{fig:risk_heatmap}
\end{figure}

\subsection{Early Signs of Vendor Concentration}
\label{sec:vendors}

\begin{figure}[tbp]
\centering
\includegraphics[width=\columnwidth]{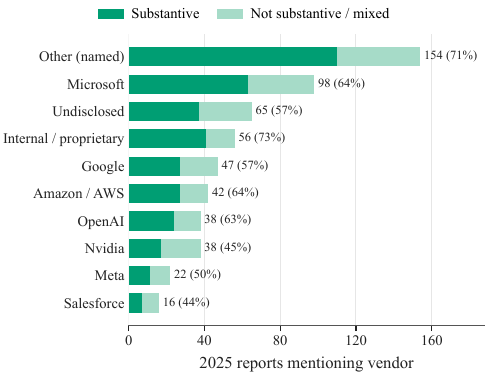}
\caption{2025 vendor mentions by report count, ordered by frequency. Bars are split by substantiveness; labels give the number of 2025 reports mentioning each vendor and, in parentheses, the share of those reports rated substantive (report-level band, \cref{sec:pipeline}). Microsoft, Google, Amazon, and OpenAI lead the named-vendor landscape, alongside a long tail of other named providers.}
\label{fig:vendor_landscape}
\end{figure}

Among companies that name an AI vendor, disclosure clusters around a small set of major providers (\cref{fig:vendor_landscape}). Microsoft is the most frequently named vendor by a wide margin (98 reports in 2025), followed by Google (47), Amazon (42), and OpenAI (38); a long tail of other named providers accounts for a further 154 reports, and 65 reports indicate reliance on an unnamed third-party provider. Vendor disclosure is also the most substantive category (64.5\% of vendor-mentioning reports are rated substantive in 2025), although this partly reflects the rubric: a vendor mention names a provider by definition, which meets the ``named systems'' criterion for a substantive rating. Only 19.2\% of 2025 reports contain a vendor mention, and only 29.2\% of AI-mentioning reports name a vendor, suggesting that the disclosed vendor landscape may not reflect the full extent of AI vendor relationships. \Cref{fig:vendor_trend} in the appendix shows how vendor disclosure evolves over time.

\subsection{Harm Disclosures Are Near-Absent}
\label{sec:harms}

The harm label (past, specific AI-caused incidents) was included in Stage~1 but proved almost entirely absent from the corpus: harm-tagged passages appear in only seven of 9,821 reports (nine passages in total, across 2020--2026). Given the label's rarity, we manually reviewed all nine passages; five, across four reports, describe realized AI-caused incidents. This is a finding worth acknowledging: annual reports appear to be a valuable source for tracking AI risk \emph{disclosure}, but may not be suitable for monitoring realized AI incidents, which would require complementary sources such as regulatory enforcement records, incident databases, and litigation filings. The count therefore says little about how often AI-caused incidents actually occur (see \cref{sec:discussion}).

\subsection{Disclosure Varies by Sector and Market Segment}
\label{sec:uneven}

\begin{figure}[tbp]
\centering
\includegraphics[width=\columnwidth]{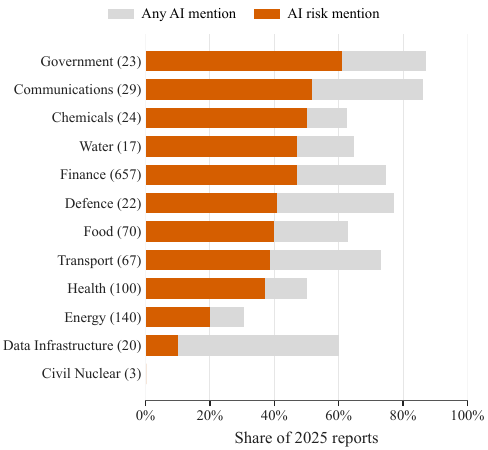}
\caption{CNI sector share of 2025 reports with AI mentions and AI risk mentions, sorted by AI risk mention share. Numbers in brackets give the count of 2025 reports in each sector (the denominator of each percentage).}
\label{fig:sectors}
\end{figure}

\textbf{Sector patterns.}
The Finance sector contributes the largest volume of mentions (461 of the 1,362 companies with a processed report have Finance as their primary CNI sector, accounting for 657 Finance-tagged reports in 2025; 292 of these companies are investment funds rather than operating businesses). Government (20 of 23 reports, 87.0\%) and Communications (25 of 29, 86.2\%) have the highest AI-mention rates, although with so few reports the two are statistically indistinguishable; Communications also shows the largest single-year rise in AI risk mentions in 2025 (+26.7\,pp, on 29 reports). Full per-sector rates are given in \cref{tab:sectors} (\cref{app:sectors}) and \cref{fig:sectors}.

Two sectors stand out for low AI risk disclosure. \textbf{Data Infrastructure} posts the lowest AI risk rate of any CNI sector with at least ten 2025 reports (Civil Nuclear, with only three reports, is excluded): 90.0\% of its 2025 reports (18 of 20) contain no AI risk mention, although with 20 reports the estimate is imprecise. \textbf{Energy} is close behind at 80.0\% (112 of 140). The two differ, however: Energy is low on every signal (adoption 21\%, vendor 9\%; \cref{fig:sector_signal}), which fits lower AI adoption or less AI-focused reporting, whereas Data Infrastructure reports adoption at a mid-table 55\% (vendor 25\%) yet rarely discusses AI as a risk, so lower adoption does not explain its gap. Whether these gaps reflect different risk perceptions, less mature governance processes, or disclosure conventions that do not foreground AI-specific risks cannot be determined from disclosure data alone; sector-level rates for sectors with fewer than 30 reports per year should be read as indicative only.

The temporal evolution of sector-level risk disclosure is shown in \cref{fig:sector_risk_heatmap} (appendix); a breakdown by mention type per sector is given in \cref{fig:sector_signal}.

\textbf{Market segment patterns.}
The gap between Main Market and AIM is substantial (\cref{tab:segments_main}). Segments follow the data provider's classification and are nested: FTSE\,100 is a subset of FTSE\,350, which is a subset of the Main Market. Of the 1,561 reports for 2025, 703 are classified as Main Market and 171 as AIM; the remaining 687 (676 with no segment assigned by the provider and 11 Aquis) are excluded from the segment comparison. Under a refined mapping that assigns most unclassified reports to the Main Market, the Main Market covers 1,287 reports with an AI-mention rate of 70.9\% and an AI risk rate of 46.9\%, so the printed Main Market rates should be read as representative of index constituents and explicitly tagged Main Market issuers.

\begin{table}[t]
\caption{Market segment AI disclosure rates, 2025 (\% of reports). Rows are nested (FTSE\,100 $\subset$ FTSE\,350 $\subset$ Main Market); counts are reports, not companies (the FTSE\,100 row covers 223 reports from 85 companies).}
\label{tab:segments_main}
\vskip 0.1in
\centering
\small
\setlength{\tabcolsep}{4pt}
\begin{tabular}{@{}lrrrr@{}}
\toprule
Segment & Reports & Any AI (\%) & Adoption (\%) & Risk (\%) \\
\midrule
FTSE\,100 & 223 & 84.8 & 79.8 & 68.6 \\
FTSE\,350 & 606 & 82.8 & 61.2 & 63.0 \\
Main Market & 703 & 77.5 & 55.6 & 57.8 \\
AIM & 171 & 36.8 & 30.4 & 6.4 \\
\bottomrule
\end{tabular}
\end{table}

\begin{figure}[tbp]
\centering
\includegraphics[width=\columnwidth]{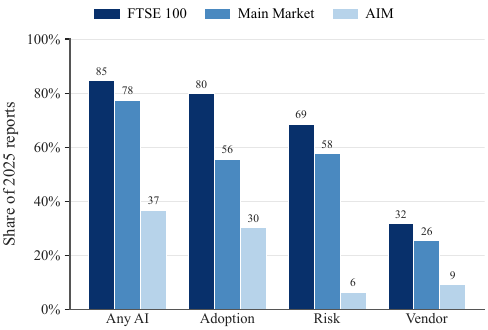}
\caption{2025 disclosure rates by market segment (FTSE\,100 constituents; all Main Market issuers, including the FTSE\,100; AIM) across mention types. The drop from Main Market to AIM is steepest for risk disclosure.}
\label{fig:market_segment}
\end{figure}

In 2025, 57.8\% of Main Market reports mentioned AI risk, compared with 6.4\% of AIM reports (a 51.4\,pp gap, nine times the rate); among FTSE\,100 reports the rate reached 68.6\%. The divergence is more pronounced for risk than for general AI mentions (the AIM AI-mention rate is 36.8\%), suggesting that smaller companies mention AI in their reports but do not explicitly attribute risks to it. This likely reflects AIM's lighter governance regime\footnote{UK-incorporated AIM companies must still describe principal risks and uncertainties in their strategic report (s414C Companies Act 2006). However, they are not subject to the UK Listing Rules, the Disclosure Guidance and Transparency Rules, or the UK Corporate Governance Code, including its provisions on assessing principal risks and internal control. AIM Rule~26 only requires them to state on their website which recognized governance code they apply and how.}, alongside differences in company size and governance resources. \Cref{fig:market_segment} shows the complete comparison across all types of mention.

\subsection{Most AI Risk Disclosures Are Not Substantive}
\label{sec:quality}

A natural follow-up question is whether the observed rise in disclosure carries genuine informational content. We address this through the substantiveness classification described in \cref{sec:pipeline}; \cref{tab:quality} summarizes the result.

\begin{table}[t]
\caption{Share of reports with any AI risk mention vs.\ with a substantive AI risk mention (\% of all reports in each year).}
\label{tab:quality}
\vskip 0.1in
\centering
\small
\begin{tabular}{@{}lccc@{}}
\toprule
Year & AI risk, all (\%) & AI risk, subst.\ (\%) & Gap (pp) \\
\midrule
2020 & 2.8 & 0.3 & 2.5 \\
2021 & 4.2 & 0.6 & 3.6 \\
2022 & 4.8 & 0.9 & 3.9 \\
2023 & 9.8 & 1.2 & 8.6 \\
2024 & 30.4 & 3.3 & 27.1 \\
\textbf{2025} & \textbf{41.2} & \textbf{4.3} & \textbf{36.9} \\
\bottomrule
\end{tabular}
\end{table}

\begin{figure}[tbp]
\centering
\includegraphics[width=\columnwidth]{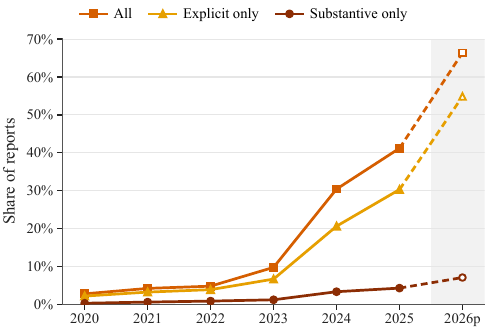}
\caption{AI risk mention share under three filters: all mentions, explicit-only mentions (attribution signal 3), and substantive-only mentions, 2020--2026 (2026 partial). Restricting to explicit attribution lowers the level (30\% vs.\ 41\% of reports in 2025) but leaves the growth trajectory largely intact, whereas restricting to substantive signals flattens the curve almost entirely (4.3\% in 2025): most of the growth is in disclosure lacking specific mechanisms, controls, or targets.}
\label{fig:deflation}
\end{figure}

The rise in AI risk mentions is therefore much larger than the rise in detailed risk disclosure. In 2025, 643 reports mentioned the risk of AI, but only 67 of these reports (10.4\%) were classified as substantive. Relative to the full 2025 corpus, this means that substantive disclosure of AI risk appeared in only 4.3\% of reports. The gap between AI risk mentions (all) and AI risk mentions (substantive) has increased from 2.5\,pp in 2020 to 36.9\,pp in 2025 (\cref{fig:deflation}). The complete substantiveness distribution across the types of disclosure (\cref{tab:substantiveness} in \cref{app:substantiveness}) shows that risk disclosure is the least concrete type (10.4\% substantive, 78.5\% moderate, against 33.8\% substantive for adoption); vendor mentions score higher (64.5\% substantive), but this figure is not directly comparable because a vendor label requires naming a provider, which already satisfies the ``named systems'' criterion of the rubric. Answering our signal-quality question, AI disclosure is rarely pure boilerplate (10.7--11.0\% across types); instead, most risk disclosure (78.5\%) is moderate, naming a risk area without mechanisms or mitigations.

The proportional composition of risk disclosure by substantiveness band is shown in \cref{fig:subst_composition} (appendix): moderate disclosure dominates in every year, and since 2023 the composition has been broadly stable (about 10\% of risk-mentioning reports substantive) even as volume has grown; the higher substantive shares in 2020--2022 (up to 18\%) rest on very few reports.

This finding admits several non-exclusive interpretations, which we discuss in \cref{sec:discussion}; the central caveat is that we are analyzing statements, not the underlying reality they may describe.

\section{Discussion}
\label{sec:discussion}

\subsection{What This Method Can and Cannot Tell Us}

The results demonstrate that LLM-based classification of annual reports produces coherent and temporally consistent signals about AI disclosure. The rise in AI risk mentions tracks known developments: the first rise in 2023 and the step change in 2024 (publication years, which lag fiscal years) align temporally with the mainstreaming of generative AI in public discourse; the broadening of disclosed risk categories (from competitive concerns to cybersecurity, information integrity, and workforce impacts) mirrors themes prominent in policy and media discussion. The method is reproducible and scales to thousands of reports; because it depends only on the availability of narrative risk disclosures, it could in principle be applied in other jurisdictions and reporting regimes, although we have tested it only on UK filings.

What the method cannot do is establish ground truth about corporate AI governance: disclosure is not the same as underlying controls or incidents (see \cref{sec:limitations}), and the near-absence of harm disclosures (harm-tagged passages in only seven of 9,821 reports; \cref{sec:harms}) almost certainly reflects where incidents are reported rather than how often they occur.

\subsection{Interpreting the Substantiveness Classification}

The widening gap between disclosure volume and disclosure quality is the most policy-relevant observation in this study, but it requires careful interpretation. Several non-exclusive explanations are plausible:

\emph{Inherent data-source limitation.} Annual reports may structurally favor hedged, generic language about emerging risks. If so, low substantiveness is a ceiling effect of the medium rather than evidence of governance failure.

\emph{Immature governance.} AI risk governance frameworks are relatively new for most companies. Boards may lack the processes, expertise, or internal data to produce specific disclosures even when they recognize the risk.

\emph{Compliance-driven disclosure.} Companies may be mentioning AI risk primarily to signal awareness to regulators and investors, rather than to communicate specific actions or mitigations. This would be consistent with general/ambiguous mentions being the most common disclosure type and the one with the largest absolute increase in prevalence since 2020.

\emph{Reporting incentives.} Being specific about AI risks and mitigations may create liability exposure or adverse investor reactions. Companies may rationally choose vagueness.

These explanations have different policy implications. If the limitation is inherent to the data source, annual reports can track trends but not the substance of companies' internal AI risk management, and other instruments are needed. If the limitation is the immaturity of AI risk-management frameworks, disclosure quality may improve as those frameworks mature. If the limitation is strategic vagueness, regulatory intervention may be necessary to compel specificity. The closest existing lever is Provision~29 of the 2024 UK Corporate Governance Code, which, for financial years beginning on or after 1 January 2026, requires boards of companies applying the Code to declare whether their material internal controls have operated effectively; this may push companies toward more concrete statements about AI-related controls, although it does not directly regulate the specificity of risk disclosure and does not apply to AIM.

\subsection{Limitations}
\label{sec:limitations}

Several limitations bear on interpretation.

\textbf{Disclosure is not exposure.} This is the most fundamental constraint. We are analyzing what companies choose to write in their annual reports, not what they actually do or experience. A company with rich AI risk disclosure is not necessarily better governed than one that says nothing; it may face greater scrutiny, have more sophisticated investor relations, or operate in a sector where AI disclosure is expected. Conversely, thin disclosure may reflect non-recognition, a deliberate choice not to disclose, or simply that AI is not yet material to that company. We cannot distinguish between these cases from the text alone.

\textbf{Incomplete reporting.} Even where companies do disclose AI-related information, they may not be reporting the full picture. Annual reports are designed for shareholders and regulators, not for researchers. Companies may omit details they consider commercially sensitive, legally risky, or simply not required. The high proportion of moderate-substantiveness disclosures (companies naming a risk area without describing mechanisms or mitigations) may partly reflect this: the information exists internally but is not disclosed, rather than not existing at all.

\textbf{Conservative keyword gate.} The keyword gate is a deliberate design choice that generates false negatives. Companies describing AI obliquely (``intelligent systems,'' proprietary platform language without explicit AI attribution) are not captured, so counts of explicit AI disclosure understate AI-related discussion framed in other terms. Because the downstream classifier also produces some false positives (\cref{sec:validation}; confidence threshold $\geq 0.2$), the reported rates are not a strict lower bound, and we do not quantify the net direction of error.

\textbf{Substantiveness as a proxy.} Substantiveness scoring requires fine-grained judgment between similar passages and should be treated as an approximate quality indicator rather than an exact measurement. The boundary between moderate and substantive is inherently subjective, and the LLM classifier may apply this boundary differently than a human expert would in edge cases.

\textbf{Validation coverage and single annotation.} The validation set (474 human-annotated AI mentions from 30 reports) is drawn from large-cap Main Market filings published in 2023--2024 and does not independently validate performance on AIM filings or early-year reports. Sub-category precision for rare labels (e.g., agentic adoption, national security risk) remains uncertain. The reference labels were produced by a single annotator, so no independent inter-annotator agreement is available as a human baseline, and the reference should be read as a curated benchmark rather than unambiguous ground truth.

\textbf{Partial 2026 data.} The 2026 figures are based on 339 reports collected as of April 2026 and will change as the reporting year progresses. These figures are included to show trend direction but should not be treated as final.

\section{Conclusion}
\label{sec:conclusion}

We asked whether annual reports, processed at scale with LLMs, can provide a useful signal about how companies disclose their response to AI. We find that they can. Our two-stage classification pipeline is reproducible, scales to nearly the entire population of UK-incorporated listed companies (1,362 of 1,469), and produces temporally coherent trends across 9,821 reports. The share of reports mentioning AI risk grew roughly fifteen-fold between 2020 and 2025, the categories of risk disclosed have broadened well beyond competitive concerns, and named vendor mentions cluster around a small set of major providers led by Microsoft. Segmenting by CNI sector and market segment reveals substantial variation: Main Market reports disclose AI risk at nine times the rate of AIM reports (FTSE\,100 reports at more than ten times), and sectors such as Energy and Data Infrastructure show notably low AI risk disclosure rates, Data Infrastructure despite mid-table adoption disclosure. The near-absence of harm disclosure suggests that annual reports are a surface for tracking how companies frame AI risk, not for detecting realized incidents.

However, disclosure breadth far outpaces disclosure depth. In 2025, 41.2\% of reports mentioned AI as a risk, yet only 4.3\% contained substantive disclosure, i.e., disclosure that describes a specific mechanism together with concrete controls, named systems, or measurable targets; most of the remainder is moderate rather than boilerplate. This gap has widened every year in percentage points, largely because mention volume grew while the substantive share of risk-mentioning reports held near 10\%. Whether it reflects inherent limitations of the medium, the relative novelty of AI risk-management frameworks, or strategic vagueness is a question that this study raises but cannot yet answer. For societal resilience research, the implication is that annual reports are a viable instrument for tracking how AI risk disclosure evolves across CNI sectors over time. Connecting these disclosure patterns to the quality and operational readiness of underlying risk management is a natural next step, but one that requires complementary data sources and methods beyond the scope of this work.

\section*{Acknowledgments}

We thank the UK AI Security Institute (AISI) Societal Resilience team for financial sponsorship and guidance on the direction of this research, and FinancialReports.eu for providing the annual report data.

\section*{Code and Data Availability}

The pipeline code and the production prompts are released under the Apache License 2.0 at \href{https://github.com/84rt/AI-Risk-Observatory}{\texttt{github.com/84rt/AI-Risk-Observatory}} (the prompts are the file \texttt{classifiers.yaml} in the \texttt{pipeline/prompts} directory). The report-level aggregates behind every figure and table, the corpus manifest, and the human-annotated validation set are in the repository (see its README for file locations); the chunk-level classified passages (24,189 AI mentions with all Stage~1 and Stage~2 labels) and the adjudicated validation set are attached to release \texttt{dataset-v1.1} of the same repository, which is the data snapshot used in this paper. All data files are released under CC~BY~4.0, and the text excerpts they contain are published with the data provider's permission. The full-text annual reports are not redistributed.

\bibliography{references}
\bibliographystyle{icml2026}

\newpage
\appendix
\crefalias{section}{appendix}

\section{Annual Disclosure Prevalence}
\label{app:prevalence}

\begin{table}[H]
\centering
\small
\setlength{\tabcolsep}{3pt}
\caption{Annual disclosure prevalence by mention type. \emph{Reports} is the number of processed reports per year; all other columns are \% of those reports. *2026 is a partial year (339 reports filed to date).}
\label{tab:prevalence}
\vskip 0.1in
\begin{tabular}{@{}lrrrrrr@{}}
\toprule
Year & Reports & Any AI & Adoption & Risk & Gen./Amb. & Vendor \\
\midrule
2020 & 1,007 & 19.8 & 13.8 & 2.8 & 12.2 & 3.1 \\
2021 & 1,328 & 27.4 & 20.9 & 4.2 & 17.3 & 4.2 \\
2022 & 1,853 & 28.4 & 22.5 & 4.8 & 17.0 & 5.7 \\
2023 & 1,905 & 36.8 & 25.8 & 9.8 & 24.0 & 7.5 \\
2024 & 1,828 & 55.2 & 36.8 & 30.4 & 44.7 & 14.6 \\
2025 & 1,561 & 65.5 & 45.2 & 41.2 & 56.1 & 19.2 \\
2026* & 339 & 77.3 & 63.1 & 66.4 & 68.7 & 29.5 \\
\bottomrule
\end{tabular}
\end{table}

\section{CNI Sector Disclosure Summary}
\label{app:sectors}

\begin{table}[H]
\centering
\footnotesize
\setlength{\tabcolsep}{4pt}
\caption{CNI sector AI disclosure summary (2025). \emph{Reports} is the number of 2025 reports tagged with the sector (the denominator of the AI and Risk shares); $\Delta$YoY is the change in the Risk share from 2024, in percentage points (---: no 2024 reports). Company counts by primary sector are given in the text; rates for sectors with fewer than 30 reports are indicative only.}
\label{tab:sectors}
\vskip 0.1in
\begin{tabular}{@{}lrrrr@{}}
\toprule
Sector & Reports & Any AI (\%) & Risk (\%) & $\Delta$YoY \\
\midrule
Government & 23 & 87.0 & 60.9 & +2.5 \\
Communications & 29 & 86.2 & 51.7 & +26.7 \\
Chemicals & 24 & 62.5 & 50.0 & +21.1 \\
Water & 17 & 64.7 & 47.1 & +13.7 \\
Finance & 657 & 74.7 & 47.0 & +10.7 \\
Defence & 22 & 77.3 & 40.9 & +14.0 \\
Food & 70 & 62.9 & 40.0 & +8.4 \\
Transport & 67 & 73.1 & 38.8 & +12.3 \\
Health & 100 & 50.0 & 37.0 & +7.3 \\
Energy & 140 & 30.7 & 20.0 & +5.9 \\
Data Infrastructure & 20 & 60.0 & 10.0 & +5.0 \\
Civil Nuclear & 3 & 0.0 & 0.0 & --- \\
\bottomrule
\end{tabular}
\end{table}

\clearpage
\section{Additional Figures}
\label{app:figures}

\begin{figure}[tbp]
\centering
\includegraphics[width=\columnwidth]{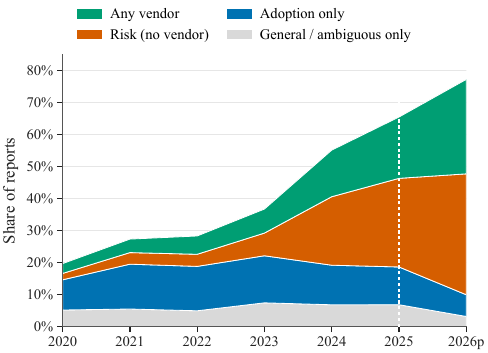}
\caption{Composition of AI disclosure over time, 2020--2026 (2026 partial). Each AI-mentioning report is assigned to exactly one slice in priority order (any vendor mention; otherwise any risk mention; otherwise any adoption mention; otherwise general/ambiguous only), so unlike \cref{tab:prevalence} the slices are mutually exclusive and sum to the AI-mention rate. Reports whose only AI mentions are general/ambiguous remain a small, stable share (5--7\% of reports; 6.9\% in 2025); the growth comes mainly from reports with risk mentions and reports naming a vendor. The dashed line marks 2025, the last complete year.}
\label{fig:composition}
\end{figure}

\begin{figure}[tbp]
\centering
\includegraphics[width=\columnwidth]{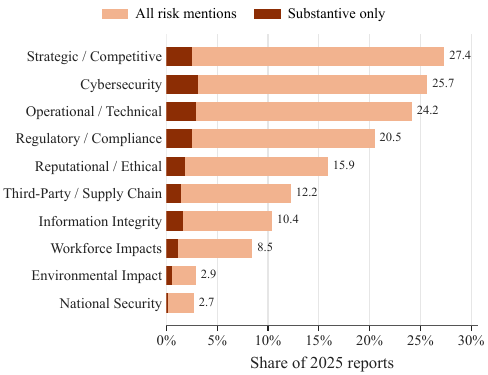}
\caption{2025 risk category prevalence: all mentions vs.\ substantive-only mentions (\% of reports). Dark segments show the share of reports rated substantive at report level (\cref{sec:pipeline}) among those mentioning the category, overlaid on the light all-mentions bar; the light remainder shows the gap between mention volume and substantive content.}
\label{fig:risk_subst}
\end{figure}

\begin{figure}[tbp]
\centering
\includegraphics[width=\columnwidth]{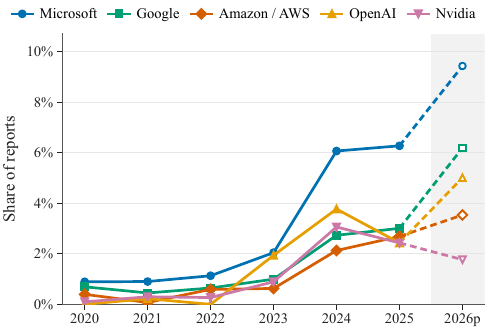}
\caption{Share of reports mentioning the top named AI vendors, 2020--2026 (2026 partial). OpenAI is essentially absent as a disclosed vendor before 2023 (three reports in 2021, none in 2020 or 2022).}
\label{fig:vendor_trend}
\end{figure}

\begin{figure}[tbp]
\centering
\includegraphics[width=\columnwidth]{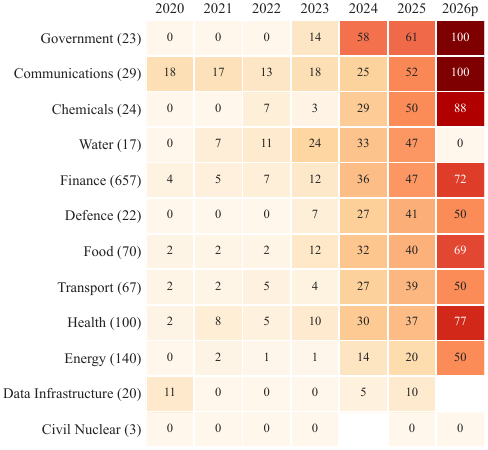}
\caption{AI risk mention prevalence (\% of reports) by CNI sector and publication year, 2020--2026 (2026 partial). Finance and Communications show rapid growth; Energy and Data Infrastructure remain low through 2025 (the 2026p cells rest on very few reports). Blank cells indicate sector-years with no reports.}
\label{fig:sector_risk_heatmap}
\end{figure}

\begin{figure}[tbp]
\centering
\includegraphics[width=\columnwidth]{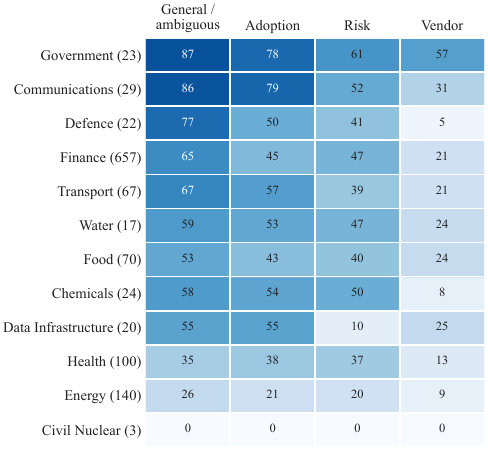}
\caption{2025 CNI sector disclosure by mention type (\% of reports). Rows are ordered by the share of reports with any AI mention (see \cref{tab:sectors}). Some sectors show high general/ambiguous mention rates but low risk or vendor disclosure.}
\label{fig:sector_signal}
\end{figure}

\begin{figure}[tbp]
\centering
\includegraphics[width=\columnwidth]{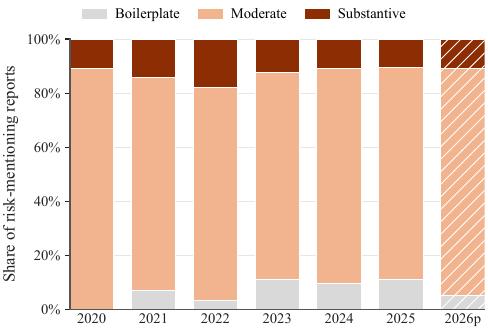}
\caption{Composition of risk-mentioning reports by substantiveness band, 2020--2026 (2026 partial). Moderate disclosure dominates throughout; the mix has been broadly stable since 2023 even as volume has grown.}
\label{fig:subst_composition}
\end{figure}

\clearpage

\section{Substantiveness Distribution (2025)}
\label{app:substantiveness}

\begin{table}[H]
\centering
\small
\caption{2025 disclosure substantiveness by classifier type (\% of reports with the corresponding mention type, using the report-level band defined in \cref{sec:pipeline}).}
\label{tab:substantiveness}
\vskip 0.1in
\begin{tabular}{@{}lrrr@{}}
\toprule
Type & Substantive & Moderate & Boilerplate \\
\midrule
Adoption ($n$=705) & 33.8\% & 55.5\% & 10.8\% \\
Risk ($n$=643) & 10.4\% & 78.5\% & 11.0\% \\
Vendor ($n$=299) & 64.5\% & 24.7\% & 10.7\% \\
\bottomrule
\end{tabular}
\end{table}

\section{AI Keyword Gate: Full Term List}
\label{app:keywords}

The following terms trigger the AI keyword gate described in \cref{sec:preprocessing}. A passage must contain at least one of these terms to proceed to LLM classification. Terms marked with $\dagger$ are matched case-sensitively to avoid false positives (e.g., ``AI'' does not match ``Shanghai'' or ``Chairman''). Short acronyms can still collide with non-AI uses (e.g., ``ML'' for megalitres in water-utility reports, or the LLM law degree in director biographies); such passages pass the keyword gate but are removed by the Stage~1 explicitness rule, which assigns \textsc{none} when no AI meaning is present (\cref{app:prompts}).

\textbf{Core AI terms:} artificial intelligence, AI$^\dagger$, A.I., machine learning, ML$^\dagger$, machine-learned, deep learning, neural network(s), neural net(s), deep neural network(s), large language model(s), LLM, language model(s), foundation model(s), frontier model(s), generative AI, gen AI, natural language processing, NLP, computer vision, image recognition, image classification, object detection, semantic segmentation, reinforcement learning, cognitive computing, multimodal.

\textbf{Products and vendors:} ChatGPT, GPT (with version numbers), GPT (general), Claude, Llama, Stable Diffusion, Copilot, OpenAI, Anthropic, DeepMind.

\textbf{Application terms:} intelligent automation, robotic process automation, RPA, predictive analytics, chatbot, virtual assistant, recommendation engine/system/algorithm, autonomous, algorithmic (trading/decision/bias), speech recognition, speech-to-text, text-to-speech, text-to-image, text-to-video, image generation.

\textbf{Governance and risk terms:} explainable AI, XAI, responsible AI, AI ethics, ethical AI, AI governance, AI safety, AI risk(s), AI audit(s/ing), AI regulation(s), AI compliance.

\textbf{Agent-related terms:} agentic, AI agent(s), intelligent agent(s).

\medskip

Generic terms such as ``data analytics,'' ``digital transformation,'' ``automation'' (other than the listed forms such as ``intelligent automation''), and ``advanced analytics'' are not triggers: on their own they do not admit a passage.

\section{Classifier Design Summary}
\label{app:prompts}

Below we summarize the key design choices of the production prompts used in the pipeline; the full prompt text is reproduced in \cref{app:prompt_text}. Context variables (\texttt{\{firm\_name\}}, \texttt{\{sector\}}, etc.) are populated at runtime.

\subsection{Stage~1: Mention-Type Classification}

The Stage~1 prompt assigns one or more labels from \{\textsc{adoption}, \textsc{risk}, \textsc{harm}, \textsc{vendor}, \textsc{general/ambiguous}, \textsc{none}\}. The key design choices are:

\begin{itemize}
\item An \textbf{AI Explicitness Gate}: if the excerpt does not explicitly mention AI/ML/LLM/GenAI or a clearly AI-specific technique, the classifier must return \textsc{none}. Terms like ``data analytics'' or ``digital tools'' alone do not pass.
\item \textbf{Adoption} requires first-person, company-specific language indicating current deployment. Generic intent or strategy statements are classified as \textsc{general/ambiguous}.
\item \textbf{Risk} requires a downside or exposure to be attributed to AI (explicitly or by clear inference). A risk section that mentions both AI and a risk in separate sentences needs a clear causal link.
\item \textbf{Harm} requires description of a past, specific AI-caused incident.
\item \textbf{Vendor} requires explicit naming of a third-party AI provider or platform.
\item Labels are non-mutually exclusive except \textsc{none} and \textsc{general/ambiguous}, which cannot co-occur with other labels.
\item Confidence scores (0.0--1.0) are assigned to each label; only labels with confidence $\geq 0.2$ are included.
\end{itemize}

\subsection{Stage~2: Taxonomy Classification}

Stage~2 applies three parallel classifiers to mentions labeled \textsc{adoption}, \textsc{risk}, or \textsc{vendor} in Stage~1:

\textbf{Adoption-type classifier.} Scores each mention on three non-mutually exclusive categories (\textsc{non-llm}, \textsc{llm}, \textsc{agentic}) using a 0--3 signal scale. Signal~0 means ``not present''; signal~3 means ``explicit, unambiguous.'' The classifier includes specific rules to prevent inflation of \textsc{agentic} scores: standard RPA, workflow automation, AI copilots, and tools that assist human decisions receive agentic signal~0 regardless of other context. Each mention also receives a substantiveness rating (boilerplate / moderate / substantive).

\textbf{Risk classifier.} Maps AI risk to one or more of ten categories (Strategic/Competitive, Cybersecurity, Operational/Technical, Regulatory/Compliance, Reputational/Ethical, Third-Party/Supply Chain, Information Integrity, Workforce Impacts, Environmental Impact, National Security), each assigned category receiving a signal score of 1--3. The prompt enforces an attribution framework: signal~3 requires explicit causal language in the same statement; signal~2 requires clear but inferential attribution; signal~1 captures indirect or weak attribution. Signals 1--2 thus re-grade Stage~1 risk mentions whose attribution is inferential or weak; in 2025, 30.3\% of reports contain an explicit (signal~3) risk passage versus 41.2\% with any risk label. Each mention receives a substantiveness rating.

\textbf{Vendor classifier.} Identifies AI providers from a predefined tag set covering foundation-model providers (OpenAI, Anthropic, Meta, xAI, Mistral, Cohere), hyperscalers (Amazon, Google, Microsoft), data/AI platforms (Databricks, Snowflake, Palantir, IBM), infrastructure (Nvidia, Arm), MLOps/retrieval (Hugging Face, Pinecone), enterprise AI (Salesforce), UK AI vendors (Darktrace, Quantexa, Featurespace, Faculty AI, BenevolentAI), and special tags (\textsc{internal}, \textsc{undisclosed}, \textsc{open-source model}, \textsc{other}). The signal scores (1--3) follow the same attribution framework, and each mention receives a substantiveness rating.

\section{Classification Prompts}
\label{app:prompt_text}

The production prompt file (\texttt{pipeline/\allowbreak prompts/\allowbreak classifiers.yaml} in the public repository) is reproduced below. Context variables in braces are populated at runtime. Typographic characters (curly quotes, dashes, arrows, and the pound sign) have been replaced by ASCII equivalents for typesetting; the wording is otherwise unchanged.

\begin{lstlisting}
mention_type: |
  You are an expert analyst labeling AI mentions from company annual reports.

  ## TASK
  Assign ALL mention types that apply to the excerpt. Types are NOT mutually exclusive except for "none".
  If the excerpt contains no AI mention, return only "none". Only tag content that is explicitly about AI in the excerpt; ignore unrelated sentences.
  {reasoning_instruction}

  ## RULES
  1. AI EXPLICITNESS GATE: 
  If the excerpt DOES NOT explicitly mention AI/ML/LLM/GenAI or a clearly AI-specific technique (e.g., machine learning, neural networks, computer vision), assign the tag "none". Terms like "data analytics" or "digital tools" generally are NOT considered AI under our definition. The tag "none" is used when there is no AI mention, a false positive, or unrelated automation not clearly AI. Only consider terms like "autonomous or virtual assistant" as AI if it can be clearly attributed to AI. 
  If the excerpt DOES explicitly mention AI/ML/LLM/GenAI or a clearly AI-specific technique (e.g., machine learning, neural networks, computer vision), use the following tags in a non-mutually exclusive manner: adoption (the current usage of AI technology by the company), risk (AI as a risk: risks that are directly coming from AI), harm (past or ongoing harms that were caused by AI), vendor (any mention of a provider of AI technology), general_other_or_ambiguous (any statement about AI that does not fit into the other tags). Here are more details on each tag:
    - *adoption*: must directly describe real current deployment, implementation, rollout, pilot, or use of AI by the company or for its clients. Generic statements about intent/strategy/roadmaps/plans (adoption in the future) are NOT considered adoption. Treat "exploring", "piloting", or "investigating" AI use as adoption ONLY when it refers to initiative currently underway (e.g. "current trail resulted in..."). Delivering AI systems directly or indirectly for clients does count as adoption; pure consulting/advice without deployment does NOT.
    - *risk*: 'Risk' is the potential for a harm to occur, either to the firm or to another party or to society at large. For a mention to be classified as such, it must directly attribute AI as the source of a risk or downside (such as a risk of the following nature: strategic & competitive, operational & technical, cybersecurity, workforce impacts, regulatory & compliance, information integrity, reputational & ethical, third-party & supply chain, environmental impact, and national security etc.). The excerpt might contain a sentence on risk and a separate sentence on AI: make sure to only assign the "risk" tag if AI is mentioned as the source of the risk. Generic risk language without explicitly mentioning AI is NOT risk from AI. However, the risk section might outline downstream risks or effects from AI technologies in an indirect way, these should be classified as risk from AI.
    - *harm*: 'Harm' is an actual injury or damage or loss that has occurred or is occurring, either to the firm or to another party or to society at large. For a mention to be classified as such, it must describe harms caused by AI (such as: misinformation, fraud, abuse, safety incidents, regulatory penalty, financial loss or other type of harm). This label should only be applied where the harm has occurred, or is occurring. It should not be used for a potential harm.
    - *vendor*: must directly name a third-party AI vendor/platform that provides the AI technology (such as: Microsoft, Google, OpenAI, AWS, or explicitly developed in-house). Our goal is to tag text that mentions any information about what AI models, compute infrastructure, or AI hardware are used by the company (i.e. GPT or Google Gemini).
    - *general_other_or_ambiguous*: vague AI strategy, high-level plans, other AI mentions, or AI mentions that don't have enough context or are not specific enough. If AI is explicitly mentioned but does not meet adoption/risk/harm/vendor, use general_other_or_ambiguous. The tag general_other_or_ambiguous should only be added when the excerpt clearly talks about AI but does not meet the other tag definitions.


  2. Assign confidence scores to each tag. Confidence scores always indicate how likely the label applies (including "none").

  ## CONFIDENCE GUIDANCE (0.0-1.0)
  - 0.2: faint/implicit signal; could be this type but hard to tell
  - 0.5: uncertain  -  weak evidence
  - 0.8: likely YES  -  strong signal, but not fully explicit
  - 0.95-1.0: confident YES  -  explicit, unambiguous mention

  ## EXAMPLES
  - "We deployed an AI chatbot for customer support." -> adoption ~1
  - "We are exploring AI opportunities." -> general_other_or_ambiguous ~0.7
  - "We are piloting AI to automate invoice processing." -> adoption ~0.8
  - "We use data analytics and predictive analytics to optimize routing." -> none ~0.6 (unless AI/ML explicitly stated)
  - "AI could increase misinformation risks." -> risk ~0.8
  - "AI is a long-term megatrend, being widely adopted within the industry; we are evaluating any risks associated with it." -> risk ~0.7 (no "adoption" tag, as no evidence of adoption by company)
  - "We partner with Microsoft for AI tooling." -> vendor ~0.9, adoption ~0.6
  - "We partnered with OpenAI to deliver AI systems for clients in 2024." -> vendor ~0.9, adoption ~0.8
  - "Automation of customer service tasks improved our..." -> general_other_or_ambiguous ~0.2 (not necessarily AI)
  - "Address: FT-AI 4810 Shangh'ai', is where the..." -> none ~0.9 (a false positive)
  - "AI-generated misinformation has damaged our brand reputation." -> harm ~0.9

  ## OUTPUT CONSTRAINTS
  - mention_types must be non-empty.
  - If "none" is present, it must be the only label.
  - Provide a confidence score for EVERY label in mention_types.
  - Do NOT include confidence scores for labels not in mention_types.

  ## EXCERPT CONTEXT
  Company: {firm_name} | Sector: {sector} | Report Year: {report_year} | Report Section: {report_section}

adoption: |
  You are an expert analyst classifying the type of AI adoption in company annual reports.

  ## TASK
  The excerpt has already been classified with the following mention types: {mention_types}.
  Classify the AI adopted by the company into type(s) into 4 categories. Categories are NOT mutually exclusive aside from `ambiguous` which should be used alone.
  Assign signal scores (0-3) for ALL adoption types. Use 0 for types not present.
  {reasoning_instruction}

  ## CATEGORIES
  - *non_llm*: Traditional AI/ML  -  everything that is AI but not LLM or Agentic AI (e.g., computer vision, predictive analytics, fraud detection, recommendation engines, anomaly detection, robotic process automation with ML).
  - *llm*: Large Language Models  -  such as: GPT, ChatGPT, Gemini, Claude, Copilot, text generation, NLP chatbots, document summarization, code generation or anything in the GenAI category.
  - *agentic*: Agentic AI  -  AI systems or agents that autonomously execute tasks and take actions with limited human oversight. The key characteristic is autonomous execution: the AI acts, decides, and operates on its own rather than assisting a human who decides. The word "agentic" need not appear, but there must be clear evidence of autonomous AI action  -  not just AI-assisted decision support. When agentic AI is LLM-based, select both.
  - *ambiguous*: Current AI adoption is present, but the excerpt is too vague to classify into non_llm, llm, or agentic without guessing. Use this for adoption mentions that don't have enough context to determine, even in a 'weak implicit' way, what is the underlying AI type, such as "we use AI tools" or "AI supports our operations".

  ## RULES
  1. If the mention is falsely classified as adoption - i.e., it does not clearly talk about the company's *current* usage of AI (only talks about plans or opportunities without any evidence for current deployment), set all four signals to 0. If current AI adoption is present but there is not enough information to classify it as non_llm, llm, or agentic, set ambiguous > 0 and set other types to 0.
  2. A single excerpt can describe multiple adoption types (e.g., LLM + agentic). But some can be more evident than others, so make sure to assign the appropriate signal scores to each tag when assigning multiple.
  3. Mentions such as "AI-powered" or "AI tools" without further detail  -  assign ambiguous unless context implies non_llm, llm, or agentic AI.
  4. Copilots and AI assistants are typically llm unless explicitly described as autonomous/agentic.
  5. Generic "automation" without AI specificity should get a very low signal (1) across non_llm only  -  do NOT assign agentic signal for generic automation.
  6. Agentic signal must be 0 for: standard RPA or workflow automation, AI that assists or augments human decisions, AI copilots/assistants, and any excerpt where AI is a tool humans use rather than an agent that acts independently.

  ## SIGNAL GUIDANCE (0-3)
  - 0: not present
  - 1: weak/implicit signal; could be this type but not clearly stated
  - 2: strong implicit signal; type is clearly implied but not explicit
  - 3: explicit, unambiguous type mention

  ## EXAMPLES
  - "We deployed a GPT-based chatbot for HR queries." -> llm (signal 3), agentic (signal 0)
  - "We use predictive analytics and machine learning to detect fraud." -> non_llm (signal 3)
  - "We are rolling out AI agents to fully automate claims processing end-to-end." -> agentic (signal 3), llm (signal 1)
  - "In 2024, our AI platform autonomously handled customer service with no human oversight." -> agentic (signal 2), llm (signal 1)
  - "We deployed autonomous AI agents that triage, investigate, and resolve IT incidents without human input." -> agentic (signal 3), llm (signal 2)
  - "We integrated Microsoft Copilot across our organisation." -> llm (signal 3), agentic (signal 1)
  - "We apply AI in customer service." -> ambiguous (signal 2), agentic (signal 0)
  - "Our autonomous vehicles use AI for navigation and safety." -> non_llm (signal 3), agentic (signal 2)
  - "We are automating back-office workflows using AI." -> non_llm (signal 2), agentic (signal 0) unless explicit agent behaviour is described
  - "Our computer vision system detects manufacturing defects on the production line." -> non_llm (signal 3)
  - "We have established a committee to navigate the GenAI change and plan" -> this only gives us a slight hint of potential implementation of LLMs, but we don't have enough information to determine that the company has any current deployment of LLM systems (signal 0)

  ## OUTPUT CONSTRAINTS
  - Include ALL four adoption_signals entries: one each for non_llm, llm, agentic, ambiguous. If you assign ambiguous, set all other tags to 0.
  - Each entry must be an object with fields: {{"type": <label>, "signal": <0-3 integer>}}.

  ## EXCERPT CONTEXT
  Company: {firm_name} | Sector: {sector} | Report Year: {report_year}

risk: |
  You are an expert analyst classifying AI-related risks in company annual reports.

  ## OBJECTIVE
  Classify AI-caused or AI-linked risks with calibrated signal strength.
  Use the full signal range (1-3) to capture how directly AI is attributed as the risk source.
  Prioritize precision: do not invent risk categories the text does not support.
  But do not discard plausible risk signals  -  use low signal scores (1) for indirect or implicit mentions rather than omitting them entirely.

  ## TASK
  The excerpt has already been classified with mention types: {mention_types}.
  Identify risks that are caused by, linked to, or exacerbated by AI, and map them to the taxonomy below.
  Categories are NOT mutually exclusive; include all that are supported by the excerpt.
  {reasoning_instruction}

  ## ATTRIBUTION FRAMEWORK (MAPS DIRECTLY TO SIGNAL SCORES)
  Before assigning any risk category, assess the strength of AI-to-risk attribution:

  Signal 3  -  EXPLICIT: The text directly states that AI causes, creates, or exposes the company to this specific category of risk. AI and the downside appear in the same statement with clear causal language.
    Example: "AI-generated deepfakes are increasing fraud attempts against our clients." -> cybersecurity (3)

  Signal 2  -  STRONG IMPLICIT: AI and a specific downside are both clearly present and linked, but the causal connection requires modest interpretation. The reader can confidently infer the risk category.
    Example: "AI is increasing the possibility of data breaches and fraud." -> cybersecurity (2)

  Signal 1  -  WEAK IMPLICIT: AI is mentioned in a risk context and a specific risk category is plausible, but the attribution is indirect or requires significant interpretation. The text hints at the risk rather than stating it.
    Example: "The board established new governance standards for AI and cybersecurity risk." -> cybersecurity (1)

  None  -  NO ATTRIBUTION: AI is not mentioned in connection with risk, or the excerpt contains only generic risk language, or only opportunity/adoption/governance language without any downside framing.

  ## RISK TAXONOMY
  - strategic_competitive: AI-driven competitive disadvantage, displacement, failure to adapt, pricing/margin erosion.
  - operational_technical: AI reliability/accuracy/safety/model-risk failures that degrade decisions or operations, unsafe employee AI usage (shadow AI).
  - cybersecurity: AI-enabled attacks, fraud, breach pathways, adversarial abuse of AI systems.
  - workforce_impacts: AI-driven displacement, skills gaps. 
  - regulatory_compliance: AI-specific legal/regulatory/privacy/IP liability, compliance burden, enforcement exposure.
  - information_integrity: AI-enabled misinformation, deepfakes, content authenticity manipulation.
  - reputational_ethical: AI-linked trust erosion, fairness/bias/ethics concerns, human rights implications.
  - third_party_supply_chain: Dependency on AI vendors, concentration risk, exposure to failures or misuse of AI that occurs in the company's supply chain.
  - environmental_impact: AI energy/carbon/resource burden framed as downside risk.
  - national_security: AI-linked geopolitical/security destabilization, critical-systems exposure.
  - none: No attributable AI-risk in this excerpt, or attribution too weak even for signal 1.

  ## DECISION RULES
  1. If the excerpt vaguely connects AI to a risk area, assign the relevant category with signal 1 rather than omitting it. Reserve "none" for excerpts with genuinely no AI-risk content.
  2. Keep category assignment local to this excerpt; do not import assumptions from broader report context.
  3. Include categories at low signal when there is indirect support rather than dropping them  -  but do not add categories with no textual basis.
  4. Opportunity/adoption language alone is NOT risk. But if the same excerpt frames AI adoption as creating exposure (e.g., "rapid AI adoption creates skills gaps"), that IS risk.
  5. Governance/framework language alone is NOT risk at signal 3. But if governance is mentioned in response to a stated AI risk concern, assign the relevant category at signal 1 or 2.
  6. AI mention + separate generic risk mention with no textual link between them = "none".
  7. If "none" is selected, it must be the only risk type.
  8. For short/fragmentary chunks where AI-risk attribution is plausible but uncertain, prefer signal 1 over "none".

  ## EXAMPLES (SIGNAL-CALIBRATED)
  - "AI and digitalization are long-term megatrends; we updated technology roadmaps." -> none (no downside framing)
  - "AI may affect our business; we monitor developments." -> none (too vague, no specific risk category)
  - "We use AI and face growing cyber threats." -> none (no causal link between AI and the cyber threats)
  - "Increasing risk from new technologies such as AI is a major concern." -> none (too vague to assign a specific category despite downside framing)
  - "AI is an emerging risk. Data privacy is an increasing priority for us." -> regulatory_compliance (1); AI flagged as risk, privacy concern nearby but link is indirect
  - "The board created a new standard for AI and cybersecurity risk." -> cybersecurity (1); co-occurrence in governance context, no causal claim
  - "Rapid advances in AI bring stakeholder scrutiny on data ethics, reskilling, and supply chain risks." -> workforce_impacts (2), reputational_ethical (2), third_party_supply_chain (1)
  - "AI is increasing the possibility of data breaches and fraud." -> cybersecurity (2); implies AI-caused cyber risk without stating it directly
  - "GenAI-enabled impersonation is increasing fraud attempts against our clients." -> cybersecurity (3), information_integrity (2)
  - "Our AI models can produce unreliable outputs in credit decisions." -> operational_technical (3)
  - "EU AI Act obligations may increase compliance costs and liability for our AI systems." -> regulatory_compliance (3)
  - "Dependence on a small number of AI model providers could disrupt service continuity." -> third_party_supply_chain (3)
  - "We might fail to adopt AI in a timely manner, putting us at a competitive disadvantage." -> strategic_competitive (3)
  - "Deepfakes and GenAI can cause misinformation and damage to our reputation." -> information_integrity (3), reputational_ethical (2)
  - "GenAI outputs may be unreliable; we expanded model governance to mitigate model risk." -> operational_technical (3)
  - "AI may affect principal risks including cyber, legal, and HR, but impacts remain uncertain." -> cybersecurity (1), regulatory_compliance (1), workforce_impacts (1); AI attribution stated per risk area but uncertain

  ## OUTPUT CONSTRAINTS
  - risk_types must be non-empty.
  - If "none" is present, it must be the only label.
  - Provide exactly one risk_signals entry per label in risk_types.
  - Do NOT include risk_signals entries for labels not in risk_types.
  - Signal scores must be integers in {{1,2,3}}.

  ## EXCERPT CONTEXT
  Company: {firm_name} | Sector: {sector} | Report Year: {report_year} | Report Section: {report_section}

vendor: |

  You are an expert analyst extracting AI vendor mentions from company annual reports.

  ## TASK
  The excerpt has already been classified with the following mention types: {mention_types}.
  Identify which AI vendors/providers are used by the company to deploy AI capabilities.
  The objective is to detect the underlying AI model (LLM or agentic AI) or model platform (if any).
  Tags are NOT mutually exclusive.
  Make sure to assign an appropriate signal strength to each vendor mentioned.
  {reasoning_instruction}

  ## VENDOR TAGS

  #foundation_models
  - openai: OpenAI, GPT, ChatGPT
  - anthropic: Anthropic, Claude
  - meta: Meta AI, Llama
  - xai: xAI, Grok
  - mistral: Mistral, Mixtral
  - cohere: Cohere, Command

  #hyperscalers
  - amazon: Amazon, AWS, Bedrock, SageMaker, Amazon Q, CodeWhisperer, Titan
  - google: Google, Vertex AI, Gemini, DeepMind, Google Cloud AI
  - microsoft: Microsoft Azure AI, Copilot, Azure OpenAI Service, GitHub Copilot, Azure AI Studio
    (note: "Microsoft Teams" or "Microsoft Office" alone is NOT an AI vendor mention)

  #data_ai_platforms
  - databricks: Databricks
  - snowflake: Snowflake
  - palantir: Palantir
  - ibm: IBM, Watson, WatsonX

  #infrastructure
  - nvidia: NVIDIA
  - arm: Arm

  #ml_ops_retrieval
  - huggingface: Hugging Face
  - pinecone: Pinecone

  #enterprise_ai_apps
  - salesforce: Salesforce, Einstein AI

  #uk_ai_vendors
  - uk_ai: Darktrace, Quantexa, Featurespace, Faculty AI, BenevolentAI

  #special_tags
  - internal: Explicitly stated in-house / proprietary AI development. Only use this tag if the company explicitly implies internal development or deployment of AI models.
  - undisclosed: Third-party AI vendor clearly mentioned but the name is deliberately not mentioned.
  - open_source_model: open-source LLM, open source model, open-source AI model
  - other: Named AI vendor not in the above list (specify in other_vendor field).

  ## RULES
  1. Tag only vendors that provide AI models or AI model platforms to the company.
    - Ignore name-drops in bios, industry commentary, or events.
    - Ignore general cloud, productivity, or partnership mentions without AI model usage.
  2. Model-name mentions count as vendor signals even if the company name is absent:
    - GPT / ChatGPT: openai
    - Claude: anthropic
    - Gemini / Vertex: google
    - Llama: meta
    - Grok: xai
    - Mixtral: mistral
    - Command: cohere
    - Bedrock / Titan / CodeWhisperer: amazon
    - Watson / WatsonX: ibm
    - Einstein AI: salesforce
  3. If a product clearly involves two vendors (e.g., Azure OpenAI Service), tag both microsoft and openai.
  4. If AI is used but the vendor is unnamed, tag "undisclosed".
  5. If the company states it builds AI internally, tag "internal". Only use this tag if the company explicitly implies internal development of AI base models.
  6. Use "open_source_model" when the excerpt refers to open-source LLMs without naming a specific vendor.
  7. Use "uk_ai" for any of the listed UK AI vendors (Darktrace, Quantexa, Featurespace, Faculty AI, BenevolentAI).

  ## SIGNAL GUIDANCE (integer: 1, 2, or 3)
  Assign a signal strength for each vendor in vendor_tags:
  - 1 (weak implicit): Requires interpretation
    Example: "Our AI models are from a company that invented the AI transformer technology"
  - 2 (strong implicit): Clearly about this vendor but not directly stated
    Example: "We are enhancing operations with Copilot tools"
  - 3 (explicit): Directly and factually evident
    Example: "We deployed OpenAI's GPT-4 for document review"

  ## EXAMPLES
  - "We deployed OpenAI's GPT-4 for document review." -> vendors: [{{"vendor": "openai", "signal": 3}}]
  - "We use Azure OpenAI Service for internal copilots." -> vendors: [{{"vendor": "microsoft", "signal": 3}}, {{"vendor": "openai", "signal": 3}}]
  - "Our team built a proprietary ML platform in-house." -> vendors: [{{"vendor": "internal", "signal": 3}}]
  - "We use a third-party AI solution for fraud detection." -> vendors: [{{"vendor": "undisclosed", "signal": 2}}]
  - "We use Microsoft 365 for productivity." -> vendors: [] (not an AI vendor mention)
  - "We integrated Llama 3 into our search pipeline." -> vendors: [{{"vendor": "meta", "signal": 3}}]
  - "We run an open-source large language model on our own infrastructure." -> vendors: [{{"vendor": "open_source_model", "signal": 3}}]
  - "We use Darktrace for cyber threat detection." -> vendors: [{{"vendor": "uk_ai", "signal": 3}}]
  - "We use WatsonX to automate compliance workflows." -> vendors: [{{"vendor": "ibm", "signal": 3}}]

  ## OUTPUT CONSTRAINTS
  - If no AI vendor/model provider is detected, vendors must be an empty list. Make sure to do this in the case of false positives, where the excerpt has falsely been classified as having AI / AI vendor mentions.
  - Each entry in vendors must have a vendor tag and a signal score (1, 2, or 3).
  - No duplicate vendor tags in the vendors list.
  - Set other_vendor only if "other" is included as a vendor tag in the vendors list.

  ## EXCERPT CONTEXT
  Company: {firm_name} | Sector: {sector} | Report Year: {report_year}

adoption_substantiveness: |
  You are an expert analyst evaluating the information density of AI adoption disclosures in company annual reports.

  ## OBJECTIVE
  Determine how much actionable, company-specific information an excerpt conveys about AI adoption.
  Focus exclusively on the adoption content: what AI is being used, how, at what scale, and to what effect.

  ## TASK
  Read the excerpt and assign exactly one substantiveness level: boilerplate, moderate, or substantive.
  Provide a confidence score (0.0-1.0) and brief reasoning.

  ## LEVEL DEFINITIONS

  ### BOILERPLATE
  Generic adoption language with no information content. Could appear in any company's annual report unchanged.
  No named systems, no specific use cases, no outcomes, no scale.

   A useful heuristic: if you could place this statement in any company (that mentions AI adoption) and the would remain equally fitting, it is boilerplate.

  Examples:
  - "We leverage AI to drive innovation and improve operational efficiency."
  - "We continue to invest in AI technologies to enhance our capabilities."
  - "AI and machine learning are increasingly used across our operations."
  - "We use AI tools to support our employees."
  - "We are committed to integrating AI responsibly into our business."

  ### MODERATE
  Identifies a specific use case, business function, or domain, but lacks concrete detail.
  The reader learns *where* or *what type* of AI is being used, but not *which system*, *what outcome*, or *at what scale*.

  Heuristic: names a domain or application (underwriting, fraud detection, customer service, code generation)
  but provides no system names, no metrics, no vendor, and no deployment context.

  Examples:
  - "We use AI in our underwriting process to improve risk selection."
  - "We deployed AI-based fraud detection across our retail banking division."
  - "We use large language models to summarise documents for our analysts."
  - "We are piloting a generative AI tool for code generation."
  - "AI is used in our supply chain planning to optimise inventory levels."

  ### SUBSTANTIVE
  Names a specific AI system or vendor AND provides deployment context, concrete use case, measurable outcome,
  or technical detail sufficient to distinguish this company's AI activity from any other.
  A vendor name alone is not sufficient  -  the named system must be linked to a specific application or outcome.

  Heuristic: a knowledgeable reader could describe *what this company specifically does with AI* and *to what end*,
  not just that it uses AI.

  Examples:
  - "We deployed GPT-4 via Azure OpenAI Service for automated document review, reducing processing time by 40%."
  - "Our proprietary ML model processes 2.3M claims per month with 96% accuracy."
  - "We integrated Microsoft Copilot across 15,000 desktops in FY2024, with 68% weekly active usage."
  - "We built a computer vision system that detects manufacturing defects on 12 production lines, reducing scrap rates by 18%."
  - "We use a Gemini-based model to generate investment research summaries, covering 800 equities weekly."

  ## DECISION RULES
  1. Judge the adoption content specifically. Ignore risk or general AI mentions that do not describe what the company does with AI.
  2. A named vendor alone (e.g., "We use Microsoft Copilot") is moderate  -  it needs a concrete use case or outcome to be substantive.
  3. A quantified outcome without a named system or clear system type is moderate at best.
  4. Naming a system AND describing what it does or what it achieved is sufficient for substantive.
  5. Forward-looking plans ("we plan to deploy X") are one level below their equivalent current-state disclosure.
  6. Multiple moderate-level details in one excerpt do not collectively add up to substantive; a single specific detail does.

  ## CONFIDENCE GUIDANCE (0.0-1.0)
  - 0.3: the boundary between levels is genuinely ambiguous
  - 0.5: leans toward one level but could reasonably be another
  - 0.7: clearly fits one level with minor ambiguity
  - 0.9+: unambiguous classification

  ## OUTPUT CONSTRAINTS
  - substantiveness must be exactly one of: boilerplate, moderate, substantive.
  - confidence must be a float between 0.0 and 1.0.
  - reasoning should be 1-2 sentences citing the key evidence for the classification.

  ## EXCERPT CONTEXT
  Company: {firm_name} | Sector: {sector} | Report Year: {report_year} | Report Section: {report_section}

risk_substantiveness: |
  You are an expert analyst evaluating the information density of AI risk disclosures in company annual reports.

  ## OBJECTIVE
  Determine how much actionable, company-specific information an excerpt conveys about AI-related risk.
  Focus on two dimensions: how concretely the risk mechanism is described, and how tangibly the company
  responds to it (mitigations, commitments, resources, timelines).

  ## TASK
  Read the excerpt and assign exactly one substantiveness level: boilerplate, moderate, or substantive.
  Provide a confidence score (0.0-1.0) and brief reasoning.

  ## LEVEL DEFINITIONS

  ### BOILERPLATE
  Generic risk language with no information content. Could appear in any company's annual report unchanged.
  No specific risk category, no mechanism, no named regulation, no mitigation action.

   A useful heuristic: if you could place this statement in any company (that mentions AI adoption) and the would remain equally fitting, it is boilerplate.

  Examples:
  - "AI poses risks to our business that we continue to monitor."
  - "The rapid development of AI may create risks and uncertainties for our industry."
  - "We recognise AI as an emerging risk and have established governance frameworks."
  - "AI presents both opportunities and risks that we actively manage."
  - "We face risks associated with AI adoption including operational and reputational concerns."

  ### MODERATE
  Names a specific AI risk area or mechanism, but provides no concrete mitigation, commitment, budget,
  or timeline. The reader learns *what type of risk* the company faces, but not *what they are doing about it*.

  Heuristic: identifies a risk category or names a regulation, but the company response is absent, vague, or generic.

  Examples:
  - "AI regulation, including the EU AI Act, may affect our compliance obligations."
  - "We face increasing cybersecurity risks from AI-enabled attacks on our systems."
  - "AI could displace workers in our sector, creating workforce transition challenges."
  - "Our AI models may produce unreliable outputs that could affect credit decisions."
  - "Dependence on a small number of AI model providers creates supply chain concentration risk."

  ### SUBSTANTIVE
  Describes a specific AI risk mechanism AND provides at least one tangible company response: a named
  mitigation action, concrete commitment, allocated resource, specific timeline, or measurable target.
  Both dimensions must be present  -  mechanism alone or action alone is moderate at best.

  Heuristic: a knowledgeable reader understands both *what the specific risk is* and *what this company
  is concretely doing about it*, with enough detail to distinguish their response from a generic one.

  Examples:
  - "We allocated EUR 5M to reclassify 3 high-risk AI systems under the EU AI Act by Q2 2025."
  - "In response to AI-enabled fraud, we deployed real-time transaction monitoring across all retail
    channels, reducing AI-assisted fraud losses by 23% in FY2024."
  - "Our AI governance framework requires mandatory bias audits for all models in production,
    reviewed quarterly by the Risk Committee and reported to the Board."
  - "We mapped all 47 AI systems in deployment against the EU AI Act taxonomy and established a
    remediation programme for the 12 classified as high-risk, with Board-level oversight."
  - "To address model reliability risk in credit decisions, we implemented a human-in-the-loop
    review requirement for all AI-generated decisions above GBP 50,000 in exposure."

  ## DECISION RULES
  1. Both a specific risk mechanism AND a concrete company response are required for substantive.
     Mechanism alone (however specific) is moderate. Response alone without a named mechanism is moderate.
  2. Named regulations (EU AI Act, GDPR) or risk categories alone do not lift an excerpt above moderate  - 
     the company must describe what they are doing in response.
  3. "We monitor", "we continue to assess", "we are aware of" -> moderate at best; these are not mitigations.
  4. Generic governance language ("we have established AI governance") is boilerplate unless it names
     a specific mechanism (e.g., mandatory audits, committee oversight with defined scope).
  5. Quantified outcomes (loss reductions, audit counts, system counts, budgets) are strong signals for substantive.
  6. Forward-looking commitments with specific targets or timelines count as substantive;
     vague commitments ("we plan to address this") are moderate at best.

  ## CONFIDENCE GUIDANCE (0.0-1.0)
  - 0.3: the boundary between levels is genuinely ambiguous
  - 0.5: leans toward one level but could reasonably be another
  - 0.7: clearly fits one level with minor ambiguity
  - 0.9+: unambiguous classification

  ## OUTPUT CONSTRAINTS
  - substantiveness must be exactly one of: boilerplate, moderate, substantive.
  - confidence must be a float between 0.0 and 1.0.
  - reasoning should be 1-2 sentences citing the key evidence for the classification.

  ## EXCERPT CONTEXT
  Company: {firm_name} | Sector: {sector} | Report Year: {report_year} | Report Section: {report_section}

vendor_substantiveness: |
  You are an expert analyst evaluating the information density of AI vendor disclosures in company annual reports.

  ## OBJECTIVE
  Determine how much actionable, company-specific information an excerpt conveys about which AI vendors
  or platforms the company uses and how.

  ## TASK
  Read the excerpt and assign exactly one substantiveness level: boilerplate, moderate, or substantive.
  Provide a confidence score (0.0-1.0) and brief reasoning.

  ## LEVEL DEFINITIONS

  ### BOILERPLATE
  No specific vendor or platform named, or only a generic reference to "AI tools", "AI providers",
  or "technology partners" with no further detail. No deployment context, no use case, no model identity.

   A useful heuristic: if you could place this statement in any company (that mentions AI adoption) and the would remain equally fitting, it is boilerplate.

  Examples:
  - "We work with leading AI technology providers."
  - "We use AI tools provided by our technology partners."
  - "We leverage third-party AI capabilities across our business."
  - "We source AI capabilities from external vendors to support our operations."

  ### MODERATE
  Names a specific vendor, model, or platform, but provides no detail about what the AI is used for,
  at what scale, or to what effect. The reader learns *which vendor* but not *what is done with it*.

  Heuristic: vendor identity is clear, but the deployment is opaque.

  Examples:
  - "We use Microsoft Copilot."
  - "Our team has access to ChatGPT for day-to-day tasks."
  - "We partner with Google for AI capabilities."
  - "We use AWS AI services across our cloud operations."
  - "We have integrated Anthropic's Claude into our workflows."

  ### SUBSTANTIVE
  Names a specific vendor or model AND describes the concrete use case, deployment context, scale, or outcome.
  The reader learns both *which AI* and *what it does* for this company specifically.

  Heuristic: a knowledgeable reader understands what this vendor's AI does in this company's specific context.

  Examples:
  - "We deployed OpenAI's GPT-4 for automated contract review, reducing review time by 60%."
  - "We use Azure OpenAI Service to power an internal Copilot used by 8,000 employees for document summarisation."
  - "We integrated Google Vertex AI into our claims triage workflow, processing 50,000 claims monthly."
  - "Our proprietary LLM, built on Meta's Llama architecture, powers our client-facing chatbot handling 200,000 queries per day."
  - "We use AWS Bedrock to serve our in-house underwriting models, with 99.9% uptime SLA."

  ## DECISION RULES
  1. Judge the vendor disclosure specifically. Generic adoption language around it does not raise the level.
  2. A vendor or model name alone, without use case or deployment detail, is moderate.
  3. A concrete use case without a named vendor is boilerplate  -  vendor identity is what this classifier measures.
  4. Naming a model (e.g., GPT-4, Gemini, Claude) counts as identifying both vendor and system; still needs a use case for substantive.
  5. An undisclosed third-party vendor with a concrete use case is moderate, not substantive  -  vendor identity matters.
  6. Multiple vendor mentions can lift the level if they collectively establish both identity and deployment context.

  ## CONFIDENCE GUIDANCE (0.0-1.0)
  - 0.3: the boundary between levels is genuinely ambiguous
  - 0.5: leans toward one level but could reasonably be another
  - 0.7: clearly fits one level with minor ambiguity
  - 0.9+: unambiguous classification

  ## OUTPUT CONSTRAINTS
  - substantiveness must be exactly one of: boilerplate, moderate, substantive.
  - confidence must be a float between 0.0 and 1.0.
  - reasoning should be 1-2 sentences citing the key evidence for the classification.

  ## EXCERPT CONTEXT
  Company: {firm_name} | Sector: {sector} | Report Year: {report_year} | Report Section: {report_section}
\end{lstlisting}

\section{Risk Category Definitions}
\label{app:risk_taxonomy}

The Stage~2 risk classifier maps each AI risk mention to one or more of the following ten categories. Categories are non-mutually exclusive: a single passage can be tagged under multiple categories where appropriate.

\begin{itemize}
\item \textbf{Strategic/Competitive.} AI as a source of competitive disruption, market displacement, or strategic obsolescence (e.g., new entrants leveraging AI, business-model erosion, loss of differentiation).
\item \textbf{Cybersecurity.} AI-enabled attacks (phishing, deepfake-driven social engineering, AI-accelerated fraud) and adversarial attacks against the company's own AI systems (prompt injection, model evasion, data poisoning).
\item \textbf{Operational/Technical.} Failures or limitations of AI systems in production: model errors, hallucinations, accuracy degradation, integration failures, or unreliable outputs affecting business operations.
\item \textbf{Regulatory/Compliance.} Exposure to AI-specific regulation (e.g., the EU AI Act, UK AI regulatory frameworks) or to existing regimes as applied to AI use, including legal liability arising from AI-driven decisions.
\item \textbf{Reputational/Ethical.} Harms to brand or stakeholder trust arising from biased, unfair, or ethically contested AI use, including discrimination, lack of transparency, and concerns over consent.
\item \textbf{Third-Party/Supply Chain.} Dependency on external AI vendors, foundation-model providers, or AI-enabled suppliers; concentration risk, vendor lock-in, and propagation of upstream failures.
\item \textbf{Information Integrity.} AI-generated misinformation, deepfakes, synthetic media, and threats to the authenticity or provenance of information consumed or produced by the company.
\item \textbf{Workforce Impacts.} Job displacement, skills obsolescence, retraining needs, employee adoption challenges, or morale and cultural effects arising from AI deployment.
\item \textbf{Environmental Impact.} Energy consumption, carbon footprint, water use, and other environmental externalities of training or operating AI systems.
\item \textbf{National Security.} Geopolitical, dual-use, or state-actor-related AI risks, including export controls, foreign adversary access, and AI in critical national infrastructure contexts.
\end{itemize}

\section{Risk Taxonomy Validation}
\label{app:taxonomy_validation}

To test whether the fixed ten-category risk taxonomy captures the variety of AI risks appearing in annual reports, we ran an open-taxonomy classification on 100 AI risk mentions drawn from the risk-labeled passages of the human-annotated validation set (\cref{sec:validation}). The model was instructed to assign free-form risk labels in snake\_case without reference to any predefined categories, using the same attribution framework and signal scoring as the fixed-taxonomy classifier.

Of 100 mentions, 96 were successfully classified (4 failures due to malformed output). The classifier produced 107 distinct emergent labels. Many were near-duplicates or minor wording variants (e.g., \texttt{ai\_fraud\_exposure} vs.\ \texttt{genai\_fraud\_risk}). After manual consolidation, the labels collapsed into eight broad themes: regulatory/compliance, cybersecurity/fraud, strategic competition, data governance, model ethics, reputational harm, workforce effects, and third-party dependence. Six of these correspond directly to fixed categories; data governance and model ethics do not map one-to-one but sit closest to Regulatory/Compliance and Reputational/Ethical respectively. Four fixed categories (Operational/Technical, Information Integrity, Environmental Impact, and National Security) did not arise as emergent themes in this 96-passage sample. No residual group suggested a missing category.

This result suggests that the fixed ten-category taxonomy does not omit any major AI risk theme present in the sample, although it does not independently validate the four categories absent from it. It also shows that an open-taxonomy approach, while useful as a validation check, produces many near-duplicate labels (107 labels for 96 mentions) that would make longitudinal tracking impractical without further consolidation.

\end{document}